\documentclass[runningheads]{llncs}

\usepackage{eccv}

\usepackage{eccvabbrv}

\usepackage{graphicx}
\usepackage{svg}
\usepackage{booktabs}
\usepackage{multirow}
\usepackage{pifont}
\usepackage{nicematrix}
\usepackage{tikz}
\usetikzlibrary{calc}

\usepackage{multicol}
\usetikzlibrary{fit,shapes.geometric}
\usepackage[table, dvipsnames, svgnames]{xcolor}

\newcommand{\cmark}{\ding{51}}
\graphicspath{ {./images/} }

\usepackage[accsupp]{axessibility}  

\usepackage{hyperref}

\usepackage{orcidlink}

\definecolor{wacvblue}{rgb}{0.21,0.49,0.74}
\definecolor{clustvitorange}{rgb}{1,0.439,0.263}
\definecolor{clustvitgreen}{rgb}{0,0.537,0.482}
\newcommand{\numberdot}[1]{%
  \protect\tikz[baseline=(char.base)]{
    \node[shape=circle, fill={brown}, text=white, inner sep=1pt] (char) {\small\textbf{#1}};
  }%
}

\begin{document}

\title{Scale Matters: Adaptive Granularity Selection for Cross-Species 3D Plant Organ Segmentation}
\titlerunning{Adaptive Granularity Selection for 3D Plant Segmentation}

\author{Carla Salazar\inst{1,2}\orcidlink{0009-0002-7437-1595} \and
Lazaros Nalpantidis\inst{1,2}\orcidlink{0000-0002-3620-4123}}
\authorrunning{C.~Salazar \and L.~Nalpantidis}

\institute{Technical University of Denmark (DTU), Kongens Lyngby, Denmark\\
\email{\{ccsna,lanalpa\}@dtu.dk}
\and Pioneer Centre for Artificial Intelligence, Copenhagen, Denmark 
}

\maketitle

\begin{figure*}
\centering
\resizebox{0.8\textwidth}{!}{
\begin{NiceTabular}{r ccc c ccc}
Plant: & \multicolumn{3}{c}{Ribes 07} & & \multicolumn{3}{c}{Rose 06} \\
\cmidrule(lr){2-4} \cmidrule(lr){6-8}
Granularity: & Scale=1 & Scale=3 & \textbf{Scale=7} & & \textbf{Scale=1} & Scale=3 & Scale=7 \\
\midrule
\textbf{Leaf} &
\begin{tabular}{@{}c@{}}\includegraphics[width=0.13\textwidth]{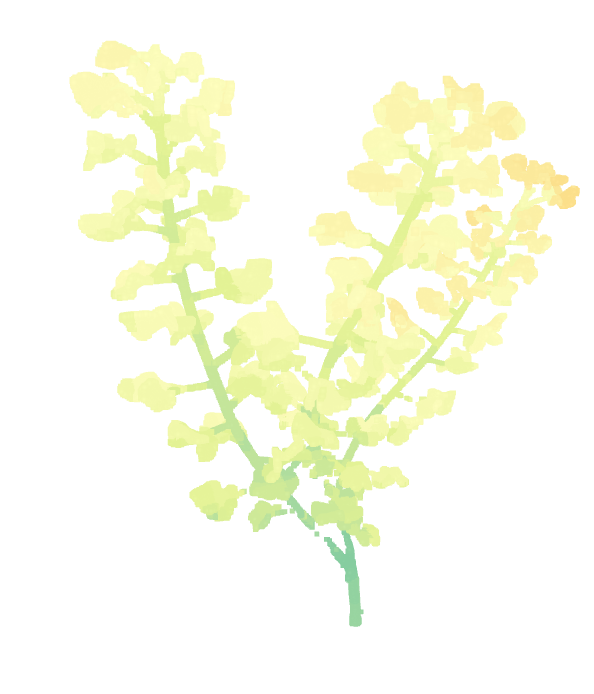}\\ \end{tabular} &
\begin{tabular}{@{}c@{}}\includegraphics[width=0.13\textwidth]{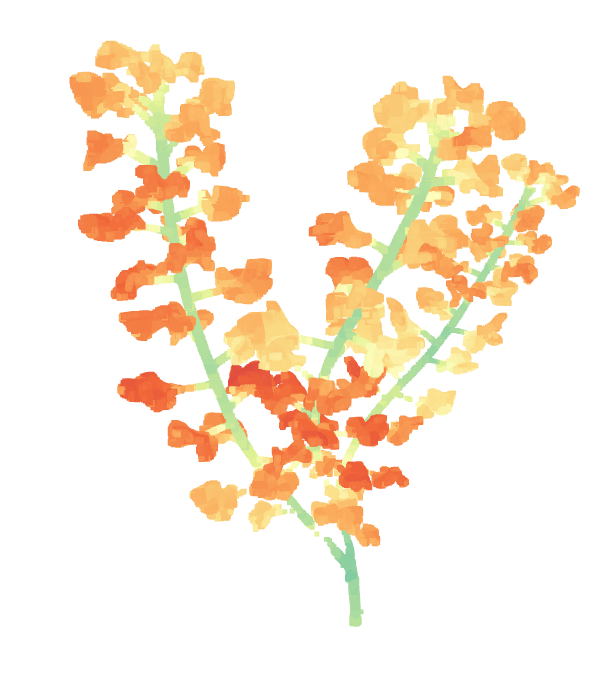}\\ \end{tabular} &
\begin{tabular}{@{}c@{}}\includegraphics[width=0.13\textwidth]{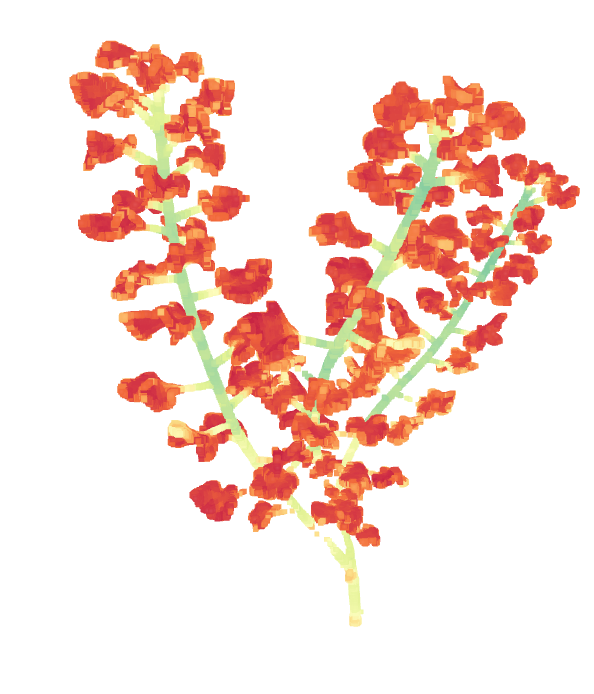}\\ \end{tabular} & & 
\begin{tabular}{@{}c@{}}\includegraphics[width=0.13\textwidth]{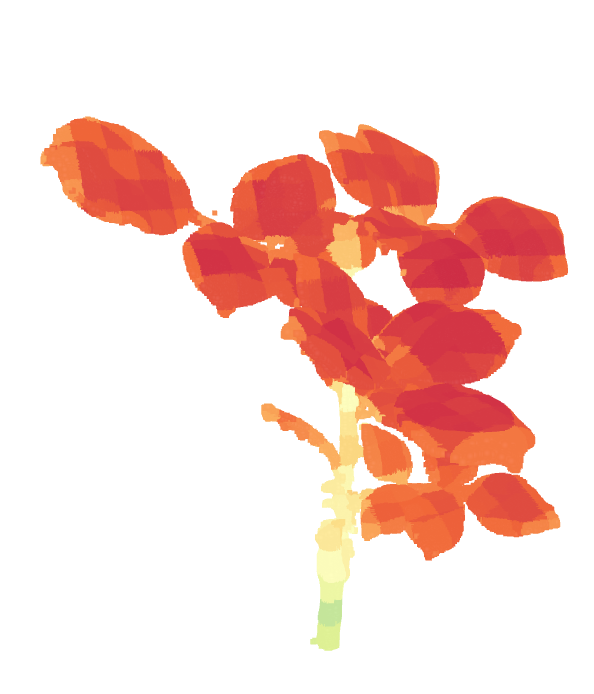}\\ \end{tabular} &
\begin{tabular}{@{}c@{}}\includegraphics[width=0.13\textwidth]{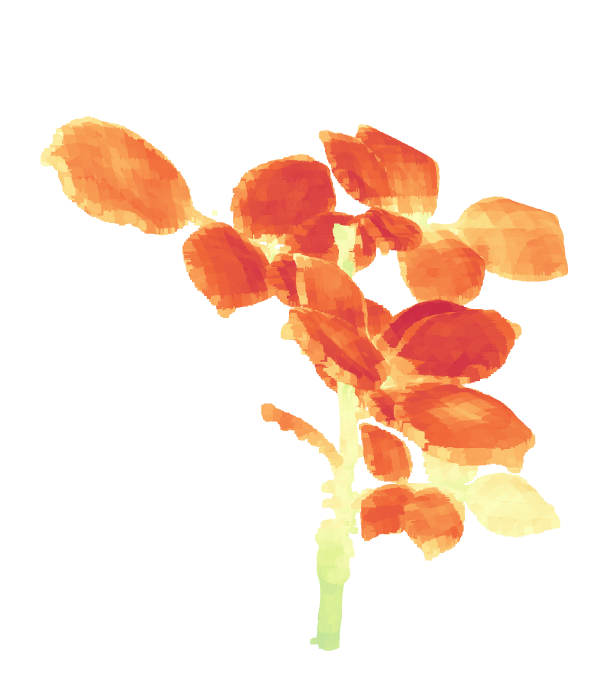}\\ \end{tabular} &
\begin{tabular}{@{}c@{}}\includegraphics[width=0.13\textwidth]{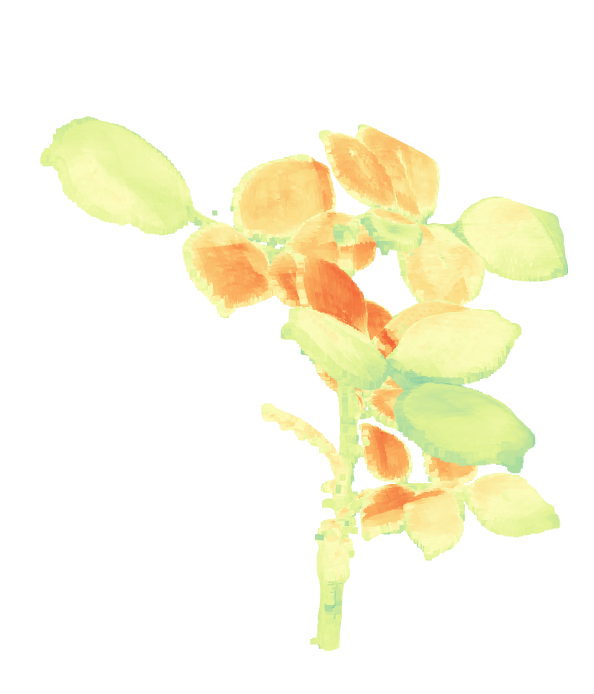}\\ \end{tabular} \\
\addlinespace
\textbf{Stem} &
\begin{tabular}{@{}c@{}}\includegraphics[width=0.13\textwidth]{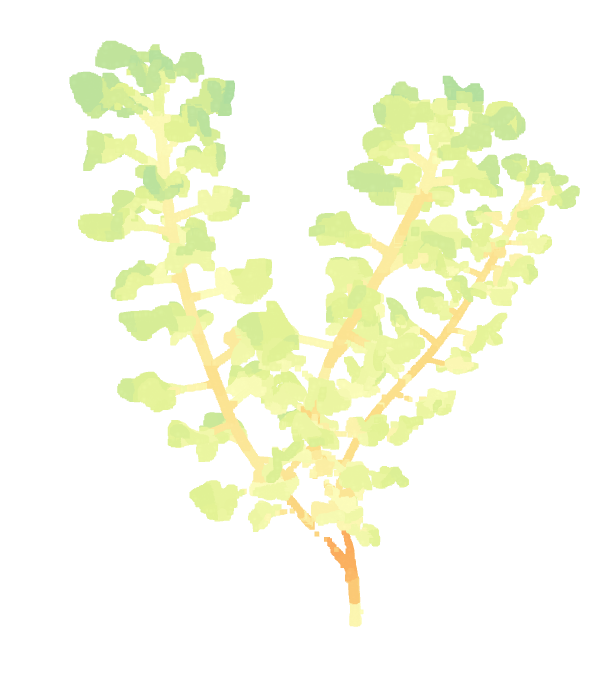}\\ \end{tabular} &
\begin{tabular}{@{}c@{}}\includegraphics[width=0.13\textwidth]{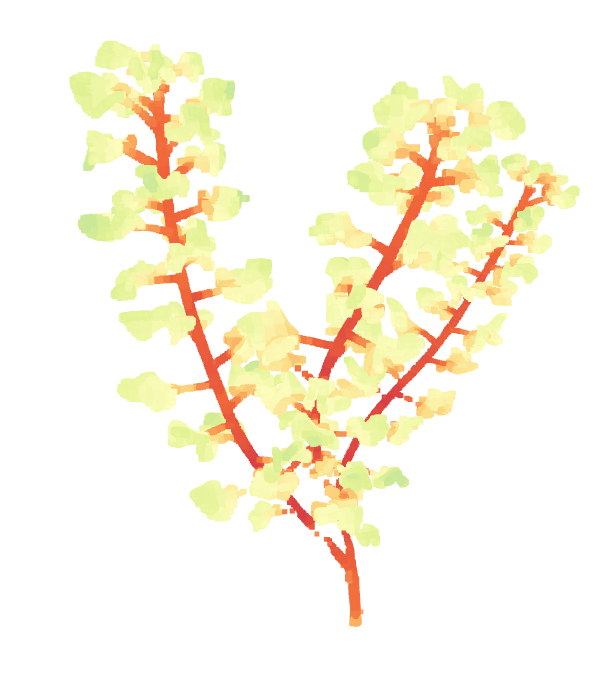}\\ \end{tabular} &
\begin{tabular}{@{}c@{}}\includegraphics[width=0.13\textwidth]{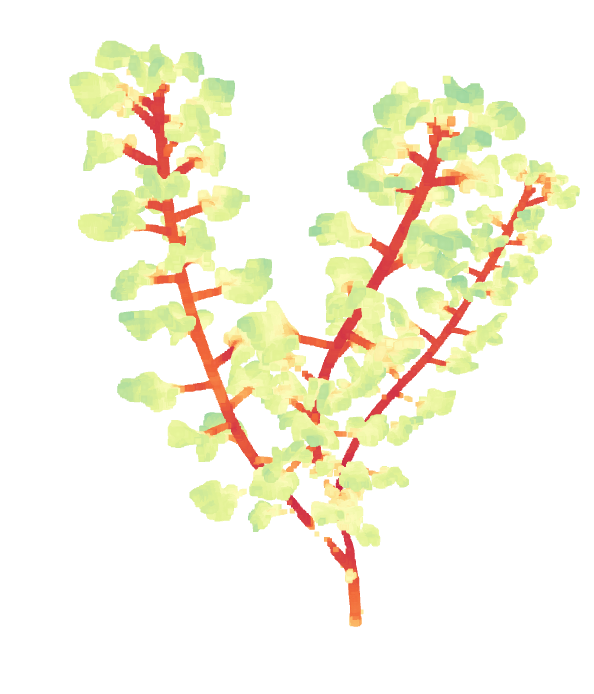}\\ \end{tabular} & & 
\begin{tabular}{@{}c@{}}\includegraphics[width=0.13\textwidth]{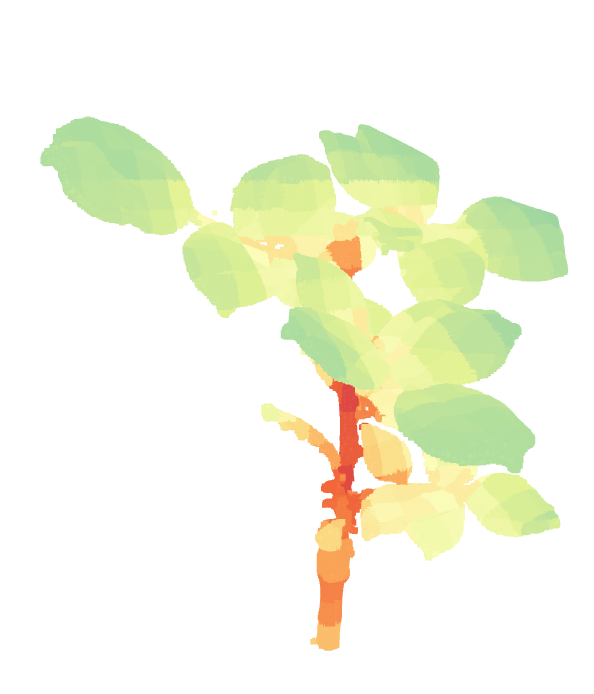}\\ \end{tabular} &
\begin{tabular}{@{}c@{}}\includegraphics[width=0.13\textwidth]{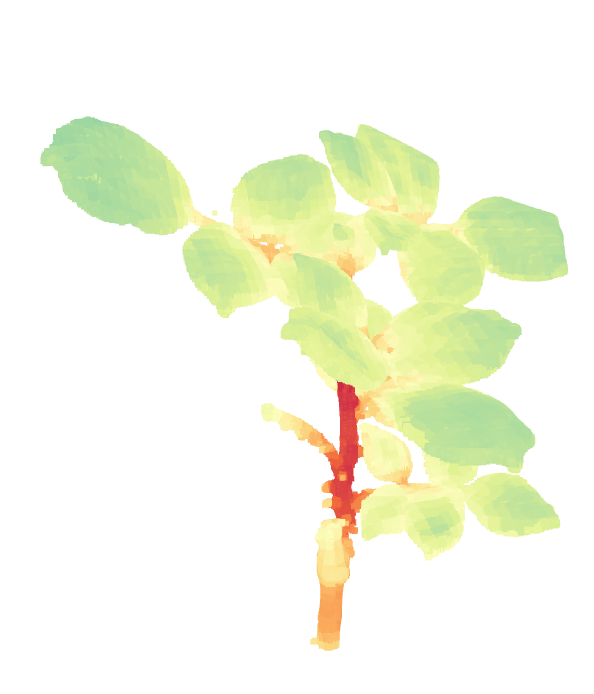}\\ \end{tabular} &
\begin{tabular}{@{}c@{}}\includegraphics[width=0.13\textwidth]{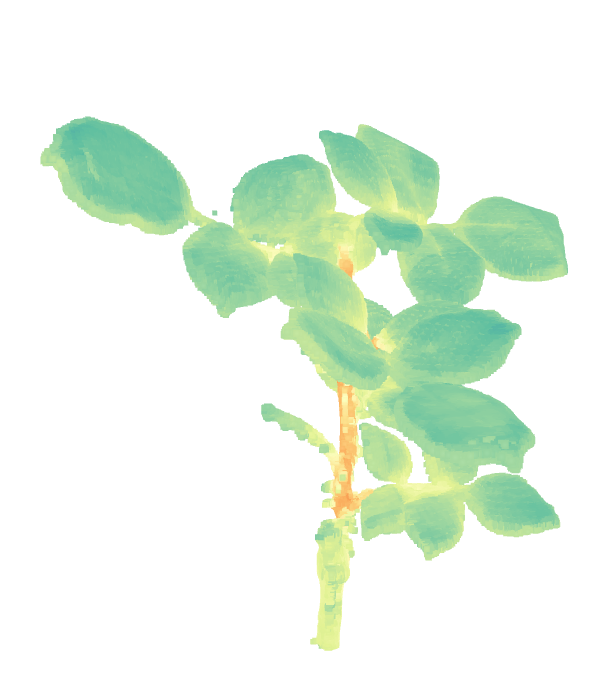}\\ \end{tabular} \\
\CodeAfter

  \tikz{
    \draw[draw=brown, line width=3pt, rounded corners]
      ($(1-1.north west)+(-33pt,7pt)$) rectangle ($(4-8.south east)+(10pt,-5pt)$);
  }
  \tikz{
    \draw[draw=gray, line width=1.5pt, rounded corners]
      ($(2-4.north west)+(-4pt,2.5pt)$) rectangle ($(4-4.south east)+(0.3pt,-0.3pt)$);
  }
  \tikz{
    \draw[draw=gray, line width=1.5pt, rounded corners]
      ($(2-6.north west)+(-4pt,2.5pt)$) rectangle ($(4-6.south east)+(0.3pt,-0.3pt)$);
  }
\end{NiceTabular}}
\caption{\textbf{The necessity of adaptive granularity in plant phenotyping.} Cosine similarity ({\color{ForestGreen}green}: low, {\color{red}red}: high) for leaves (top) and stems (bottom) between prototypes and Utonia features across different input scales for two example plants. A single fixed granularity fails to generalize across varying plant morphologies. The first plant requires a much finer granularity (scale=7) compared to the second (scale=1).}
\label{fig:scale_similarity}
\vspace{-25px}
\end{figure*}

\begin{abstract}
Recent 3D foundation models provide powerful feature representations for point cloud learning by controlling spatial granularity. However, relying on a fixed spatial granularity severely limits generalization in applications like plant phenotyping, where organ morphology and size vary substantially across species and growth stages. To address this, we propose AGS-PlantSeg, a few-shot 3D plant organ segmentation method that leverages the frozen Utonia \cite{utonia2026} foundation model combined with Adaptive Granularity Selection. By dynamically selecting the best granularity levels for each specific plant model, our method extracts optimized geometric features for a lightweight MLP segmentation head. Extensive experiments across PLANesT-3D \cite{planest3d2025}, Pheno4D \cite{pheno4d2021}, and Crops3D \cite{crops3d2024} demonstrate that AGS-PlantSeg significantly improves cross-species generalization, achieving 88.9\% average mIoU performance and outperforming fixed-granularity baselines by 2.5 mIoU points. Despite requiring minimal annotated data, our approach is highly competitive with fully supervised, plant-specific architectures. Project page: \url{https://github.com/DTU-PAS/ags-plantseg}
\end{abstract}

\section{Introduction}
\label{sec:intro}

Computer vision is increasingly used for plant phenotyping as a tool to measure traits such as leaf area, fruit count, and organ morphology \cite{cvforplants}. A key requirement for these measurements is plant organ segmentation, where organs such as leaves, stems, flowers, and fruits must be identified within a plant image or 3D scan. While 2D image-based methods are limited by occlusions and missing spatial information \cite{rosesegnet2022}, 3D point cloud-based approaches provide geometric cues that motivate recent work on 3D plant organ segmentation \cite{rosesegnet2022,crops3d2024,planest3d2025}.

Recent 3D foundation models have transformed point cloud learning by enabling strong feature representations across tasks and domains. Models such as Concerto \cite{concerto2025} and Utonia \cite{utonia2026} have shown that pretrained 3D encoders can generalize beyond a single dataset, moving seamlessly from robotic manipulation to large-scale outdoor scenes. In Utonia, this generalization is supported by the control of spatial granularity at which the model represents local structures. In many tasks, this granularity can be selected before deployment as the expected object size, scene scale, or point cloud density is relatively consistent. 
 
However, this reliance on consistent spatial granularity presents a significant challenge in plant phenotyping, where organ size and morphology vary substantially across datasets, species, growth stages, and even between organs within the same plant. As a result, manually selecting a single fixed granularity is non-trivial: the granularity level that accurately represents stems and leaves for one species may not be suitable for another, especially when organs differ in size, density, or spatial arrangement. Similarly, even within the same species, the appropriate scale may change as plant organs shift in proportion across growth stages. As different granularities produce different feature representations, relying on a fixed choice limits generalization when plant morphology changes.
Fig.~\ref{fig:scale_similarity} illustrates this effect by comparing the cosine similarity between Utonia features and learned stem and leaf prototypes across two distinct example plants: Ribes 07 and Rose 06. The preferred granularity differs significantly between the two, demonstrating that a plant's specific morphology directly affects which granularity level best captures the spatial boundary between stem and leaf regions.

To address this limitation, we introduce \emph{AGS-PlantSeg}, a plant organ segmentation method that combines the Utonia foundation model with Adaptive Granularity Selection. Instead of training a plant-specific segmentation network from scratch or manually choosing a fixed granularity level, we keep the Utonia encoder frozen and allow AGS-PlantSeg to automatically select the feature levels that best match each individual plant model. These selected levels are then used to extract frozen features that accurately reflect the morphology of the target plant. Finally, a lightweight MLP-based segmentation head operates on these optimized representations to classify plant organs. 
These design choices, endow AGS-PlantSeg with the following properties:
\numberdot{1} It remains competitive with fully supervised baselines while using frozen 3D foundation model features,
\numberdot{2} it can transfer to unseen species and datasets, and finally
\numberdot{3} its Adaptive Granularity Selection module improves segmentation performance compared with fixed-granularity alternatives.

The main contributions of this paper are:
\begin{itemize}
    \item We propose AGS-PlantSeg, a few-shot 3D plant organ segmentation method that combines frozen 3D foundation model features with an Adaptive Granularity Selection (AGS) module.
    \item We introduce a prototype-based granularity scoring strategy that selects suitable feature granularity levels during both training and inference, utilizing metrics for organ separation, within-organ compactness, and boundary consistency.
    \item We demonstrate that adaptive granularity selection improves cross-species generalization for plant organ segmentation, achieving performance competitive with fully supervised, plant-specific baselines without requiring retraining for every new plant model.
\end{itemize}

\section{Related Work}
\label{sec:related}

\smallskip \noindent \textbf{3D Segmentation.} 
3D point cloud segmentation has evolved from early point-based architectures such as PointNet \cite{pointnet2016} and PointNet++ \cite{pointnet++2017} to more recent architectures with improved local feature aggregation, such as PointNeXt \cite{pointnext2022}, and attention-based designs, such as Point Transformer variants \cite{wu2024ppt,wu2022ptv2,wu2024ptv3}.
These developments have also influenced 3D plant organ segmentation, where several methods train dedicated networks to segment organs such as leaves and stems at the semantic or instance level.
For example, PSegNet \cite{psegnet2022} incorporates attention modules within a deep learning neural network for both semantic and instance segmentation. GCASSN \cite{gcassn2025} combines graph convolutional network with self-attention to improve feature learning on plant point clouds. 
Other methods build on established point cloud backbones such as PointNet++ \cite{pointnet++2017} or PointNeXt \cite{pointnext2022}. For example, Dong et al. \cite{automaticseg2025} combine PointNeXt with Quickshift++ to increase the generalization ability of the model. 
These methods achieve strong performance in supervised settings, and some report improved generalization across plant species \cite{automaticseg2025,gcassn2025}. 
However, generalization remains challenging because plant morphology, organ size, and point cloud structure can vary substantially across species, growth stages, and acquisition settings.\\

\noindent \textbf{3D Foundation Models.}
Recent work has started to explore pretrained 3D point cloud encoders that can provide transferable representations across different tasks and domains.
Models such as Sonata \cite{sonata2025} and Concerto \cite{concerto2025} show that self-supervised pretraining can produce strong 3D representations for downstream tasks, reducing the need to learn task-specific features entirely from scratch.
Utonia \cite{utonia2026} extends this further, demonstrating cross-domain and cross-scale generalization within a single model.
Such pretrained encoders are of particular interest for 3D plant organ segmentation, where datasets are scarce and often small, and where morphology varies substantially across species, growth stages, and acquisition settings.
However, using frozen 3D foundation models for plant organ segmentation introduces an additional challenge: the desired feature granularity may depend on the target structure.
This issue is related to recent work in 3D part segmentation, where several methods have explored controllable or promptable segmentation granularity for object part segmentation \cite{sampart2024,partfield2025,s2sam3d2026}.
These works indicate that part-level segmentation is not only a question of feature quality, but also of selecting an appropriate level of detail for the target task.
\section{Methodology}

\begin{figure}[t]
    \centering
    \includegraphics[width=\linewidth]{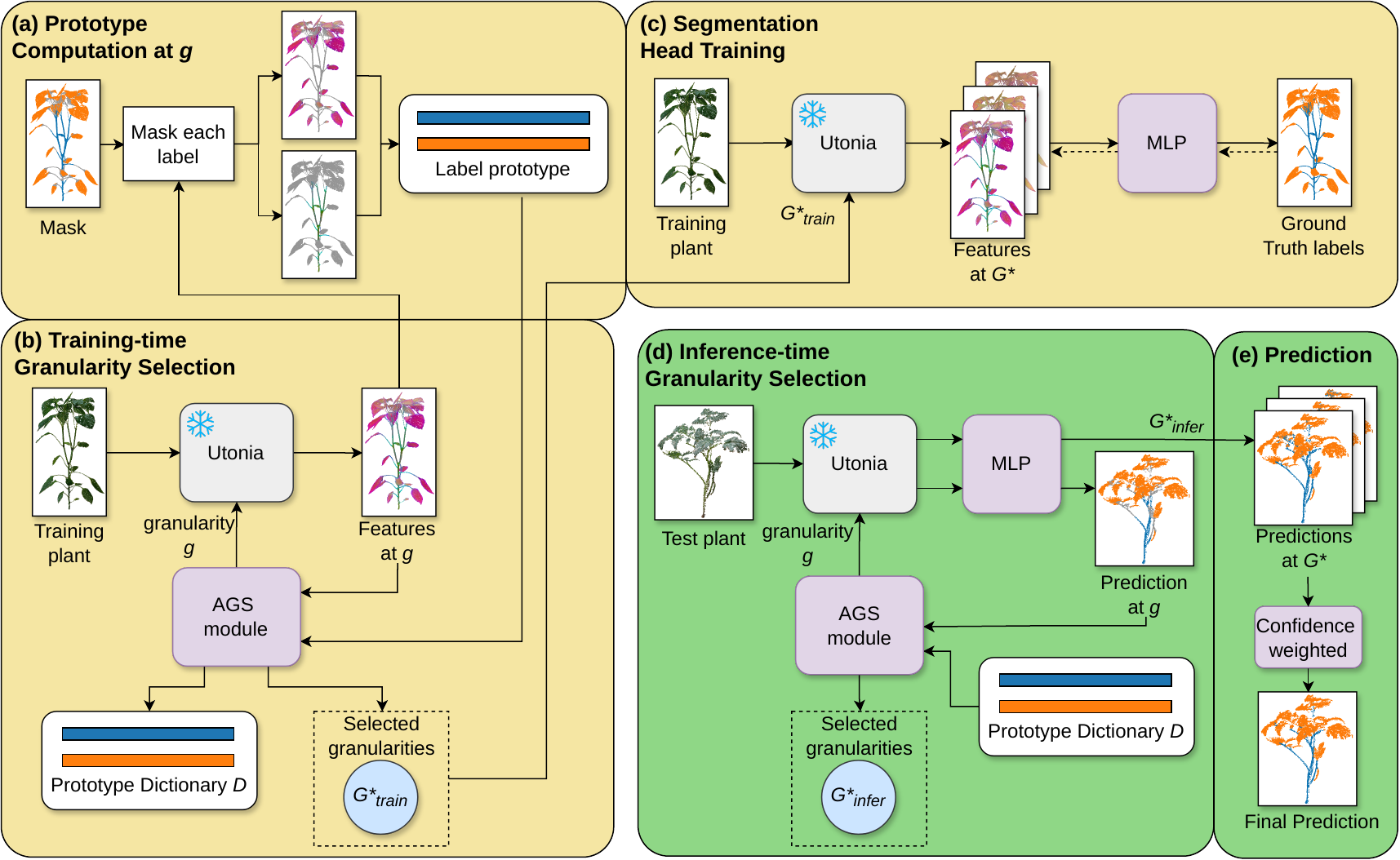}
    \caption{\textbf{Overview of AGS-PlantSeg.} The method consists of five stages: (a) prototype computation at a candidate granularity level $g$, (b) training-time granularity selection using prototype-based scores, (c) segmentation head training on the selected granularity levels $G_{\mathrm{train}}^*$, (d) inference-time granularity selection using stored training prototypes and predicted labels, and (e) final prediction from the selected granularity levels $G_{\mathrm{infer}}^*$. Orange and blue denote the two semantic organ labels. Feature spaces are visualized with PCA for illustration.}
    \label{fig:methodology}
    \vspace{-0pt}
\end{figure}

The proposed AGS-PlantSeg selects the feature granularity that is most suitable for each plant scenario, where a scenario may differ in acquisition setup, plant species, growth stage, or organ morphology. Instead of assuming that one fixed granularity is optimal for all plants, we evaluate how well each granularity represents the semantic structure needed for organ segmentation.

Intuitively, a suitable granularity should satisfy three criteria. First, features belonging to different organs should be well separated, so that stem and leaf regions occupy distinct areas of the feature space. Second, clusters of features belonging to the same organ should be compact, so that points from the same class remain close to each other. Third, transition regions between organs should be treated carefully, since points near stem-leaf boundaries may contain mixed geometric or semantic information. 

Granularity levels are selected during both training and inference using the Adaptive Granularity Selection (AGS) module of AGS-PlantSeg. The selection is implemented through a prototype-based scoring system, inspired by the use of prototypes in few-shot learning \cite{prototypes2017}. During training, labeled plants are evaluated at different candidate granularity levels by computing boundary-filtered prototypes and scoring how well each granularity separates organs, preserves compact organ representations, and handles boundary regions. The selected granularity levels are then used to train the segmentation head. 
During inference-time granularity selection, ground-truth labels are not available, so AGS-PlantSeg uses predicted labels from the trained segmentation head to compare test-plant features against the stored training prototype dictionary and select suitable granularity levels. This full process is summarized in Fig.~\ref{fig:methodology}.

\vspace{5pt} \noindent \textbf{Utonia Features and Boundary-filtered Prototypes.}\\
We build our method on features from the frozen Utonia model \cite{utonia2026}. Given a plant point cloud and a granularity level $g$, Utonia extracts a feature vector $f_i^{g}$ for each point $i$. Changing $g$ changes the granularity of the resulting feature representation, allowing the model to emphasize different spatial structures of the plant.

To summarize each semantic organ in this feature space, we use class prototyping \cite{prototypes2017}. A prototype is the average feature vector of all points belonging to a given class. Since the feature representation depends on the granularity level, prototypes are also computed separately for each $g$. For example, the stem prototype at granularity level $g$ is obtained by averaging the Utonia feature vectors of stem points extracted at $g$, while the leaf prototype is obtained from leaf points. These prototypes are later used to evaluate how well a candidate granularity level $g$ separates organ features, preserves compact class representations, and handles boundary regions.

However, not all labeled points are equally reliable for defining an organ prototype. Points near the boundary between two organs may contain mixed information, especially around stem--leaf transitions. Including these points in the prototype computation can shift the prototype away from the interior representation of the organ. Therefore, before computing prototypes, we remove boundary points from the labeled plant.

A point is considered a boundary point if at least one of its $K$ nearest neighbors belongs to a different semantic class. Only the remaining interior points are used to compute the class prototypes. This produces boundary-filtered prototypes that better represent the central feature distribution of each organ. In Fig.~\ref{fig:methodology}(a), this corresponds to applying the boundary mask, masking each label separately, and computing one prototype per semantic class at candidate granularity level $g$.

\vspace{5pt} \noindent \textbf{Training-time Granularity Selection.}\\
During the training-time granularity selection stage, corresponding to Fig.~\ref{fig:methodology}(b), ground-truth labels are available, so each candidate granularity level $g$ can be evaluated directly against the criteria introduced above. The first criterion requires features from different organs to be well separated. We measure this using the inter-class distance between class prototypes:

\begin{equation}
\label{eq:Dinter}
D_{\mathrm{inter}}^g =
\frac{1}{|\mathcal{P}|}
\sum_{(c,c') \in \mathcal{P}}
d_{\cos}\left(p_{c}^{g},p_{c'}^g\right),
\end{equation}
where $\mathcal{P}$ is the set of class pairs, $p_c^g$ is the prototype of class $c$ at granularity level $g$, and $d_{\cos}$ denotes cosine distance.

The second criterion requires clusters of features from the same organ to remain compact. We measure this using the intra-class distance between interior point features and their corresponding class prototype:
\begin{equation}
\label{eq:Dintra}
D_{\mathrm{intra}}^g =
\frac{1}{|\mathcal{C}|}
\sum_{c \in \mathcal{C}}
\frac{1}{|\mathcal{I}_c|}
\sum_{i \in \mathcal{I}_c}
d_{\cos}\left(f_i^g,p_c^g\right),
\end{equation}
where $\mathcal{C}$ is the set of semantic classes, $\mathcal{I}_c$ is the set of interior points with label $c$, and $f_i^g$ is the feature vector of point $i$ at granularity level $g$.

A good granularity level should maximize inter-class distance while minimizing intra-class distance. Inspired by Fisher-style class separability \cite{fisher1936}, where discriminative representations should have high between-class separation and low within-class variation, we define the training discriminability score as:
\begin{equation}
\label{eq:Sdisc}
S_{\mathrm{disc}}^g =
\frac{D_{\mathrm{inter}}^g}
{D_{\mathrm{intra}}^g+\epsilon},
\end{equation}
where $\epsilon$ is a small constant for numerical stability. Higher values of $S_{\mathrm{disc}}^g$ indicate that the feature space better separates organs while preserving compact organ representations.

The third criterion concerns boundary regions, where features may contain mixed information from neighboring organs. To evaluate whether these points remain consistent with their semantic class, we also compute an intra-class boundary distance from boundary points to their corresponding class prototype:

\begin{equation}
\label{eq:Dboundary}
D_{\mathrm{boundary}}^{g,c}
=
\frac{1}{|\mathcal{B}_{c}|}
\sum_{i \in \mathcal{B}_{c}}
d_{\cos}\left(f_{i}^{g}, p_{c}^{g}\right)
\end{equation}

Here, $\mathcal{B}_c$ is the set of boundary points with ground-truth label $c$. Since lower distances indicate stronger consistency with the class prototype, we define the boundary score as:

\begin{equation}
\label{eq:Sboundary}
S_{\mathrm{boundary}}^{g,c}
=
1 - D_{\mathrm{boundary}}^{g,c}
\end{equation}

Thus, in our binary setting with classes $A$ and $B$, each candidate granularity level is assigned three training-time scores. AGS selects the top-ranked granularity for each score. If multiple scores select the same granularity, it is retained only once.  
The corresponding class prototypes are stored in a training prototype dictionary $\mathcal{D}$, which is later used during inference:

\begin{equation}
\label{eq:trainingscores}
\mathbf{s}_{\mathrm{train}}^g =
\left[
S_{\mathrm{disc}}^g, \quad
S_{\mathrm{boundary}}^{g,A}, \quad
S_{\mathrm{boundary}}^{g,B}
\right]
\end{equation}

\begin{equation}
\label{eq:Gtrain}
G_{\mathrm{train}}^*
=
\left\{
g_{\mathrm{disc}}^*, \quad
g_{A}^*, \quad
g_{B}^*
\right\}.
\end{equation}

\vspace{5pt} \noindent \textbf{Segmentation Head Training.}\\
In the segmentation head training stage, corresponding to Fig.~\ref{fig:methodology}(c), Utonia features are extracted at the selected training granularity levels $G_{\mathrm{train}}^*$. 
Each point-wise feature vector extracted at a selected granularity level is treated as an individual training sample for the shared MLP, paired with the corresponding ground-truth organ label. We keep the segmentation head small so that the evaluation focuses on the selected Utonia features and the AGS module, rather than on model capacity.

\vspace{5pt} \noindent \textbf{Inference-time Granularity Selection.}\\
The inference-time scoring procedure corresponds to Fig.~\ref{fig:methodology}(d). At inference time, ground-truth labels are not available. Therefore, AGS-PlantSeg cannot compute ground-truth prototypes for the test plant or directly evaluate inter-class separation as in training.
Instead, candidate granularity levels are evaluated against a stored training prototype dictionary. This dictionary contains the class prototypes computed from the training plants and summarizes the feature distributions on which the segmentation head was trained.
We therefore select granularity levels that produce test-plant features similar to this training feature structure.

At inference time, the stored prototype dictionary $\mathcal{D}$ is used to evaluate candidate granularity levels. Each entry $d \in \mathcal{D}$ contains the class prototypes $\tilde{p}_{d,c}$ from one training plant. For a candidate granularity level $g$, the segmentation head first predicts labels $\hat{y}_i^g$ for each point. These predicted labels are used to define predicted interior points $\hat{\mathcal{I}}_c^g$ and predicted boundary points $\hat{\mathcal{B}}_c^g$ for each class $c$.

To approximate the intra-class compactness criterion at inference time, we measure how close predicted interior-point features are to the stored training prototypes. Since the test plant does not need to match every training plant, we compute this distance against each stored prototype pair and keep the best match.

\begin{equation}
\label{eq:dinterior}
D_{\mathrm{interior}}^{g,d} = \frac{1}{|\mathcal{C}|}
\sum_{c \in \mathcal{C}}
\frac{1}{|\hat{\mathcal{I}}_c^g|}
\sum_{i \in \hat{\mathcal{I}}_c^g}
d_{\cos}\left(f_i^g,\tilde{p}_{d,c}\right).
\end{equation}

\begin{equation}
\label{eq:Sinterior}
S_{\mathrm{interior}}^{g} = \max_{d \in \mathcal{D}}
\left(1 - D_{\mathrm{interior}}^{g,d}\right).
\end{equation}

Higher values of $S_{\mathrm{interior}}^{g}$ indicate that the candidate granularity level $g$ produces interior features that are close to at least one stored training representation. This is used as an inference-time proxy for selecting a granularity level compatible with the feature space on which the segmentation head was trained.

Following the same boundary-consistency criterion used during training, we evaluate whether predicted boundary points remain close to the prototype dictionary for their predicted class. 
For each class $c$, we compare predicted boundary-point features to the corresponding class prototypes in the dictionary and keep the best matching score:

\begin{equation}
\label{eq:InfDboundary}
D_{\mathrm{boundary}}^{g,c,d} = \frac{1}{|\hat{\mathcal{B}}_c^g|}
\sum_{i \in \hat{\mathcal{B}}_c^g}
d_{\cos}\left(f_i^g,\tilde{p}_{d,c}\right).
\end{equation}

\begin{equation}
\label{eq:InfSboundary}
S_{\mathrm{boundary}}^{g,c} = \max_{d \in \mathcal{D}}
\left(1 - D_{\mathrm{boundary}}^{g,c,d}\right).
\end{equation}
Thus, each candidate granularity level is assigned three inference-time scores. 
As in training-time, AGS selects the top-ranked granularity level for each score:

\begin{equation}
\label{eq:Sinference}
\mathbf{s}_{\mathrm{infer}}^g =
\left[
S_{\mathrm{interior}}^{g}, \quad
S_{\mathrm{boundary}}^{g,A}, \quad
S_{\mathrm{boundary}}^{g,B}
\right].
\end{equation}

\begin{equation}
\label{eq:Gtest}
G_{\mathrm{infer}}^* =
\left\{
g_{\mathrm{interior}}^*, \quad
g_A^*, \quad
g_B^*
\right\}.
\end{equation}

\vspace{5pt} \noindent \textbf{Prediction.}\\
As shown in Fig.~\ref{fig:methodology}(e), the trained MLP is applied to Utonia features extracted at the selected inference granularity levels $G_{\mathrm{infer}}^*$. 
For each point $i$, the MLP outputs a class probability distribution $p_{i,g,c}$ for each selected granularity level $g$ and class $c$:
\begin{equation}
\label{eq:prediction}
\hat{y}_i = \arg\max_c \sum_{g \in G_{\mathrm{infer}}^*} w_{i,g} p_{i,g,c} ,
\end{equation}
where $w_{i,g}$ is the normalized confidence weight for point $i$ at granularity level $g$, computed from the maximum predicted class probability at that level.

\section{Experimental Setup}
\label{sec:experimental_setup}
\subsection{Dataset}

In this paper, we use three public datasets: PLANesT-3D \cite{planest3d2025}, Crops3D \cite{crops3d2024} and Pheno4D \cite{pheno4d2021}. Our approach has only been trained on PLANesT-3D data, using Crops3D and Pheno4D exclusively as test sets to evaluate generalization of our model to unseen species and datasets. This protocol is applied across all experiments and ablations.

The PLANesT-3D \cite{planest3d2025} is formed by point clouds of 10 pepper plants, 14 ribes plants and 10 rose plants, annotated by leaf and stem. This dataset provides a confidence score per point and, following the original protocol, points with confidence below 6 are removed. 
We evaluate three training settings. The full-split is the official PLANesT-3D split, where pepper (01, 03, 07), ribes (03, 10, 11, 14), and rose (01, 03, 09) are used for testing, and the remaining are used for training, corresponding to approximately a 70-30 split. In the few-shot setting, we chose one plant of each species for training: pepper 02, ribes 02 and rose 02. In the one-shot setting, we train only on pepper 02 and test on the remaining pepper plants as well as all ribes and rose plants. 

From Crops3D \cite{crops3d2024}, we use the 3D scans of tomato and potato plants as an additional test set. The dataset contains 83 tomato and 118 potato plants annotated with stem and leaf labels. For the tomato subset, we exclude 10 scans that include fruit annotations, as the training data does not contain a fruit class.

From Pheno4D \cite{pheno4d2021}, we used their 77 annotated 3D point clouds of 7 tomato plants scanned daily during a 3 week period. This dataset is annotated by soil, stem and leaf, however, we remove the soil data for these experiments as the training data only contains stem and leaf.

\subsection{Implementation Details}

All point clouds are downsampled to 200K points to reduce memory consumption. During feature extraction, Utonia is kept frozen and features are computed at the granularity levels selected by our Adaptive Granularity Selection module.
Following the official Utonia recommendation for object-level inputs \cite{utonia-github}, we apply coordinate normalization and a scale transformation.
Coordinate normalization is used in all experiments to bring point clouds from different datasets into a comparable coordinate range, since the original scans may differ substantially in units or spatial scale. The scale transform is then used as the granularity parameter searched by our Adaptive Granularity Selection.

The granularity search is initialized at a scale of 4.0 and evaluates candidate scales with a step size of 0.5 within the range [1.0, 12.0]. Candidate scales are ranked using the proposed scoring system. We compute the prototypes excluding boundary points, where boundary detection is performed using a $k$-nearest-neighbor filter with $k=1000$.

The extracted multi-scale features are used to train a MLP-based lightweight segmentation head. For each training plant, the point-wise features are split into 80\% training and 20\% validation sets using stratified sampling to preserve the class distribution. For the full-split setting, storing features for every point would exceed GPU memory. We therefore randomly sample 50K points from each 200K-point cloud after Utonia feature extraction, reducing memory requirements. For few-shot and one-shot splits, we maintain all extracted points without subsampling.

The segmentation head is a two-layer MLP with a 256-dimensional hidden layer, trained on frozen Utonia features using Adam with a learning rate of $10^{-3}$, batch size $2048$, and cross-entropy loss. Training runs for at most $200$ epochs with early stopping after $15$ epochs without validation improvement, and uses dropout of $0.3$. All experiments were conducted on a single NVIDIA GeForce RTX 5090.

\subsection{Evaluation Metrics}

We report mean Intersection over Union (mIoU) and Accuracy (Acc). The mIoU is computed as the average IoU over all semantic classes, while Accuracy is computed as the fraction of correctly classified points over the total number of evaluated points.

\section{Experiments and Results}

We conduct experiments to support our claimed properties of AGS-PlantSeg, as stated in Sec.~\ref{sec:intro}:
\numberdot{1} We first show that AGS-PlantSeg remains competitive with fully supervised baselines while using frozen 3D foundation model features. 
\numberdot{2} We then evaluate its transfer to unseen species and datasets. 
\numberdot{3} Finally, we show that Adaptive Granularity Selection improves segmentation performance compared with fixed-granularity alternatives.

\subsection[Baseline Comparison 1]{Baseline Comparison \numberdot{1}}

We compare AGS-PlantSeg against PointNet++~\cite{pointnet++2017}, RoseSegNet~\cite{rosesegnet2022}, and SP-LSCnet~\cite{planest3d2025}, as reported on the official PLANesT-3D split~\cite{planest3d2025}. GCASSN~\cite{gcassn2025} uses a different 80/20 split and preprocessing protocol and is included for reference. Since source code is unavailable, we report published results.

Table \ref{tab:baseline-comparison} shows that in the full-split setting, our method achieves a mIoU of 96.3\%. This matches the best reported average mIoU while obtaining the highest average accuracy among the compared methods. In particular, our method performs strongly on pepper and ribes, reaching 96.9\% and 97.5\% mIoU, respectively.
The few-shot setting remains close to the full-split setting. When trained with only one plant per species, our method obtains 94.9\% average mIoU, compared with 96.3\% in the full-split setting. This suggests that the frozen Utonia features provide a strong representation for plant organ segmentation even with limited annotated data.
In the one-shot setting, performance remains high for pepper, which is included in training, and for ribes, which is unseen during training. This indicates that the frozen Utonia features can support cross-species generalization with limited supervision. However, performance drops on rose, reducing the average mIoU to 93.3\%. Since rose is a more challenging morphology, this suggests that additional fine-tuning may still be needed for difficult species.

\begin{table}[t]
\centering
\caption{\textbf{Comparison with baseline methods on PLANesT-3D (\%)}. PointNet++, RoseSegNet, and SP-LSCnet are reported from PLANesT-3D~\cite{planest3d2025}, using the same 70/30 plant-level split as our full setting. GCASSN~\cite{gcassn2025} uses a different 80/20 split and preprocessing protocol. Best result is shown in bold and second-best is underlined.}
\label{tab:baseline-comparison}
    \resizebox{\textwidth}{!}{
\begin{tabular}{lccccccccccccccccc}
\toprule
\multirow{2}{*}{Method} && \multirow{2}{*}{Supervision} && \multicolumn{2}{c}{Pepper} && \multicolumn{2}{c}{Rose} && \multicolumn{2}{c}{Ribes} && \multicolumn{2}{c}{Average} && \multirow{2}{*}{\shortstack{Rel.\\Training\\Speed}}  \\ 
\cmidrule(lr){5-6} \cmidrule(lr){8-9} \cmidrule(lr){11-12} \cmidrule(lr){14-15}
 &&  && Acc & mIoU && Acc & mIoU && Acc & mIoU && Acc & mIoU \\
\midrule
PointNet++ \cite{planest3d2025}  && Full && 97.9 & 94.3 && 96.7 & 89.8 && 98.4 & 94.2 && 97.7 & 92.8 && -\\
RoseSegNet \cite{planest3d2025}  && Full && 98.3 & 95.5 && 97.4 & 92.0 && 98.6 & 95.1 && 98.1 & 94.2 && -\\
SP-LSCnet \cite{planest3d2025}   && Full && 98.1 & 95.0 && 97.1 & 89.8 && 98.3 & 94.5 && 97.8 & 93.1 && -\\
GCASSN \cite{gcassn2025}   && Full && 95.6 & \textbf{97.1} && 92.8 & \textbf{95.4} && 93.7 & \underline{96.5} && 94.0 & \textbf{96.3} && -\\
\midrule
\rowcolor{gray!15}
Ours        && Full    && 98.8 & 96.9 && \textbf{98.5} & \underline{94.4} && \textbf{99.2} & \textbf{97.5} && \textbf{98.8} & \textbf{96.3} && 1.0$\times$\\ 
\rowcolor{gray!15}
Ours        && Few-Shot && \textbf{99.0} & \underline{97.0} && \underline{97.9} & 92.6 && \underline{99.0} & 95.2 && \underline{98.6} & \underline{94.9} && 2.0$\times$\\
\rowcolor{gray!15}
Ours        && One-Shot && \underline{98.9} & 96.9 && 96.8 & 88.0 && \underline{99.0} & 95.0 && 98.2 & 93.3  && 6.0$\times$ \\
\bottomrule
\end{tabular}
}
\end{table}

\subsection[Generalization to Unseen Plant Species and Datasets 2]{Generalization to Unseen Plant Species and Datasets \numberdot{2}}

\begin{figure*}[t]
\centering
\setlength{\tabcolsep}{2pt}
\renewcommand{\arraystretch}{0.5}
\begin{tabular}{cccc}
& Point Cloud & Prediction & Ground Truth \\

\rotatebox{90}{\parbox{2.2cm}{\centering Tomato\\(Pheno4D)}} &
\includegraphics[width=0.22\linewidth]{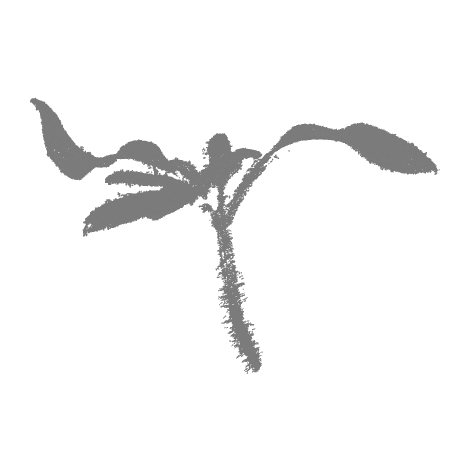} &
\includegraphics[width=0.22\linewidth]{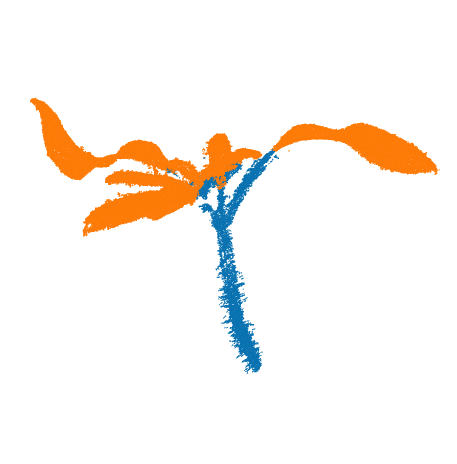} &
\includegraphics[width=0.22\linewidth]{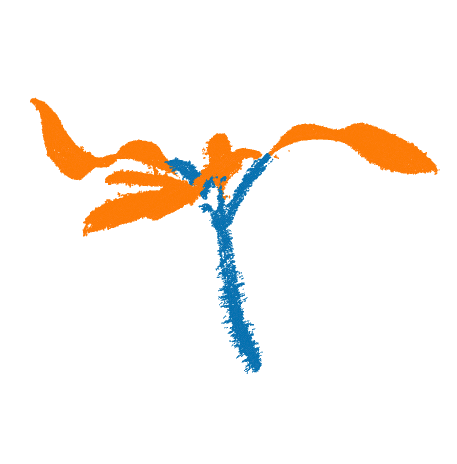} \\

\rotatebox{90}{\parbox{2.2cm}{\centering Tomato\\(Crops3D)}} &
\includegraphics[width=0.22\linewidth]{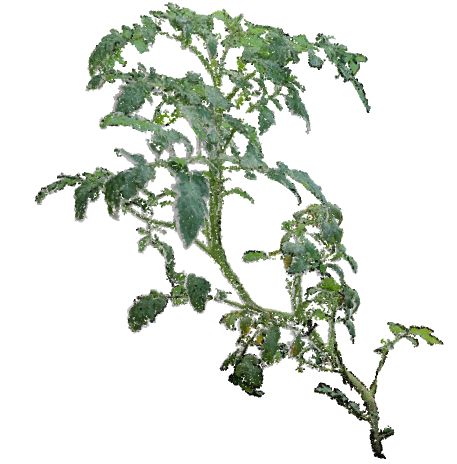} &
\includegraphics[width=0.22\linewidth]{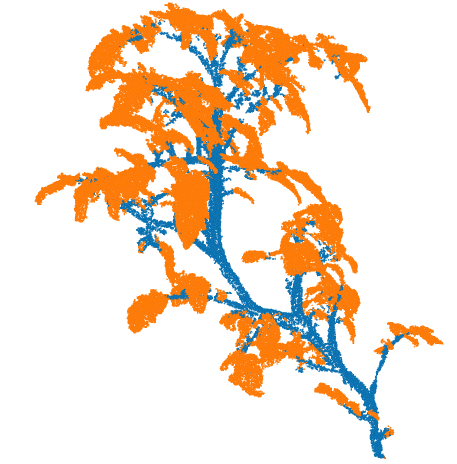} &
\includegraphics[width=0.22\linewidth]{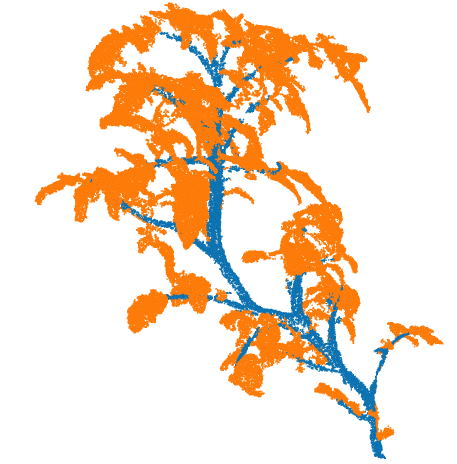} \\

\rotatebox{90}{\parbox{2.2cm}{\centering Potato\\(Crops3D)}} &
\includegraphics[width=0.22\linewidth]{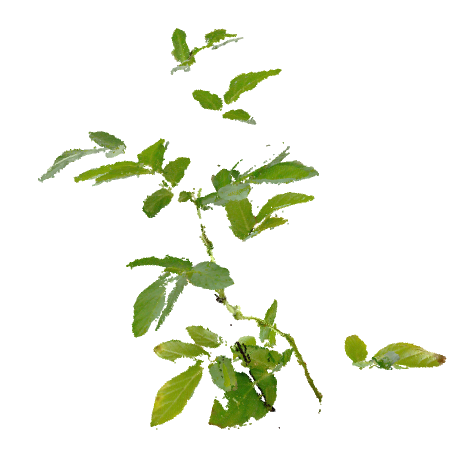} &
\includegraphics[width=0.22\linewidth]{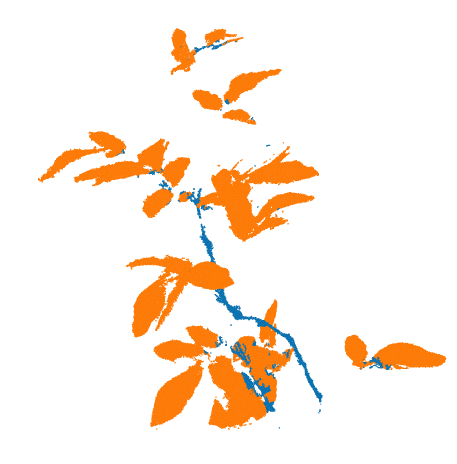} &
\includegraphics[width=0.22\linewidth]{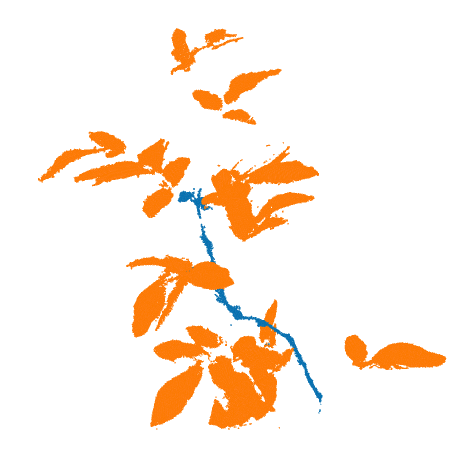} \\

\end{tabular}
\caption{\textbf{Qualitative generalization results on unseen species and datasets.} Predictions are shown alongside ground truth for Pheno4D tomato and Crops3D tomato and potato. Orange denotes leaves and blue denotes stems.}
\label{fig:qualitative_generalization}
\end{figure*}

Figure~\ref{fig:qualitative_generalization} and Table~\ref{tab:generalization} show the generalization performance on unseen species and datasets. Overall, the proposed method transfers well beyond the plants used for supervision. The full, few-shot, and one-shot models achieve comparable average performance, reaching 84.7\%, 82.9\%, and 82.5\% mIoU, respectively. This suggests that the frozen Utonia features already provide a transferable representation for organ segmentation, so only limited annotated data is needed to adapt the classifier to the task.

\begin{table}[t]
\centering
\caption{\textbf{Generalization to unseen plant species and datasets (\%)}. AGS-PlantSeg is evaluated with full, few-shot, and one-shot supervision. Baseline results are reported as published in the corresponding papers.
Best result is shown in bold and second-best is underlined.}
\label{tab:generalization}
\begin{tabular}{lccccccccccccc}
\toprule
\multirow{3}{*}{Method} && \multirow{3}{*}{Supervision} && \multicolumn{2}{c}{Pheno4D} && \multicolumn{4}{c}{Crops3D} && \multicolumn{2}{c}{\multirow{2}{*}{Avg}} \\
\cmidrule(lr){5-6} \cmidrule(lr){8-11}
 &&  && \multicolumn{2}{c}{Tomato} && \multicolumn{2}{c}{Tomato} & \multicolumn{2}{c}{Potato} \\
\cmidrule(lr){5-6} \cmidrule(lr){8-9} \cmidrule(lr){10-11} \cmidrule(lr){13-14}
 &&  && Acc & mIoU && Acc & mIoU & Acc & mIoU && Acc & mIoU \\
\midrule
GCASSN \cite{gcassn2025}   && Full  && 91.5 & 60.8 && - & - & - & - && - & - \\
\midrule
\rowcolor{gray!15}
Ours   && Full    && \textbf{97.1} & \textbf{86.0} && \textbf{94.3} & \textbf{82.6} & 97.2 & 85.6 && \textbf{96.2} & \textbf{84.7} \\
\rowcolor{gray!15}
Ours && Few-Shot && 96.6 & 83.7 && 93.6 & 81.3 & 96.7 & 83.6 && 95.6 & 82.9 \\
\rowcolor{gray!15}
Ours   && One-Shot && 94.0 & 81.4 && 93.2 & 80.2 & \textbf{97.5} & \textbf{85.9} && 94.9 & 82.5 \\
\bottomrule
\end{tabular}
\end{table}

On Pheno4D tomato, GCASSN reported 60.8\% mIoU when trained on pepper and tested on tomato. In comparison, our full-split model reaches 86.0\% mIoU, and the few-shot model reaches 83.7\% mIoU. This suggests that frozen foundation features, combined with scale selection, provide a strong cross-species generalization.
Interestingly, the one-shot model achieves the highest performance on potato with an mIoU of 85.9\%, despite being trained on a single pepper plant. However, this result is not uniform across plant species: performance is strong on Ribes and Crops3D potato, but lower on Rose and tomato plants. This suggests that one-shot supervision can work very well when the target morphology is close to the training example, but is less stable across larger morphological shifts.

Qualitatively, Figure~\ref{fig:qualitative_generalization} shows predictions and ground-truth labels for Pheno4D tomato and Crops3D tomato and potato. While some local errors remain visible, the predicted organ structure is largely preserved across unseen datasets.

\subsection[Ablations 3]{Ablations \numberdot{3}}

We conduct ablation studies in the few-shot setting to analyze the contribution of the AGS module. The ablations are divided into two parts. First, we compare AGS against fixed-granularity alternatives, where Utonia features are extracted at a single fixed granularity for both training and inference. These results are shown in Table~\ref{tab:fixed_granularity}. Second, we evaluate different components of AGS, including boundary filtering, granularity scoring, and confidence-weighted prediction. These results are shown in Table~\ref{tab:ags_components}.

\vspace{5pt} \noindent \textbf{Fixed-granularity Alternatives}\\
We compare AGS against the common fixed-granularity setting, where the same MLP segmentation head is trained using Utonia features extracted at a single scale transform. 
We evaluate seven fixed scales in Table~\ref{tab:fixed_granularity}, ranging from 1.0 to 7.0, to sample the granularity range densely enough to capture the variation in plant morphology across species and datasets. This comparison isolates the effect of adaptive granularity selection from the segmentation architecture.

In contrast, AGS does not commit to a single global granularity. While it does not achieve the best mIoU for every individual species, it ranks first or second across all evaluated species and datasets. This consistency leads to the highest average performance, with 88.9\% mIoU. The results show that adaptive granularity selection improves robustness over fixed-granularity feature extraction, while the cases where a fixed granularity still performs better indicate that our granularity selection strategy could be further improved.

\begin{table}[t]
\centering
\caption{\textbf{Comparison with fixed-granularity alternatives in the few-shot setting.} Fixed baselines use a single Utonia scale for both training and inference, while AGS selects granularity levels adaptively. Best result is shown in bold and second-best is underlined.}
\label{tab:fixed_granularity}
\resizebox{\textwidth}{!}{
\begin{tabular}{ccccccccccccccccccc}  
\toprule
\multirow{3}{*}{Scale} && \multicolumn{6}{c}{PLANesT-3D} && \multicolumn{2}{c}{Pheno4D} && \multicolumn{4}{c}{Crops3D} && \multicolumn{2}{c}{\multirow{2}{*}{Avg}} \\
\cmidrule(lr){3-8} \cmidrule(lr){10-11} \cmidrule(lr){13-16}
 && \multicolumn{2}{c}{Pepper} & \multicolumn{2}{c}{Rose} & \multicolumn{2}{c}{Ribes} && \multicolumn{2}{c}{Tomato} && \multicolumn{2}{c}{Tomato} & \multicolumn{2}{c}{Potato} \\
\cmidrule(lr){2-4} \cmidrule(lr){5-6} \cmidrule(lr){7-8} \cmidrule(lr){10-11} \cmidrule(lr){13-14} \cmidrule(lr){15-16} \cmidrule(lr){17-19}
&& Acc & mIoU & Acc & mIoU & Acc & mIoU && Acc & mIoU && Acc & mIoU & Acc & mIoU && Acc & mIoU \\
\midrule
1.0 && 97.4 & 92.7 & 96.1 & 86.7 & 97.3 & 87.7 && 93.8 & 70.6 && 92.0 & 77.2 & \textbf{97.1} & \textbf{84.8} && 95.6 & 83.3 \\
2.0 && 98.3 & 95.2 & 97.5 & 91.0 & 98.3 & 92.3 && 89.5 & 74.8 && \textbf{93.8} & \underline{81.2} & 95.7 & 79.0 && 95.5 & 85.6 \\
3.0 && 98.7 & 96.4 & 97.4 & 90.7 & 98.8 & 94.5 && 95.1 & \underline{83.4} && 91.1 & 76.4 & 94.3 & 77.2 && \underline{95.9} & \underline{86.4} \\
4.0 && \underline{98.8} & \underline{96.8} & 97.6 & 91.5 & \underline{98.9} & 94.9 && \underline{95.6} & 83.0 && 91.1 & 76.5 & 92.0 & 71.3 && 95.7 & 85.7 \\
5.0 && \underline{98.8} & 96.7 & \textbf{98.0} & \textbf{93.0} & 98.6 & 93.3 && 91.7 & 75.3 && 88.5 & 72.2 & 90.4 & 68.2 && 94.3 & 83.1 \\
6.0 && \textbf{98.9} & \textbf{97.0} & 97.8 & 91.8 & \textbf{99.1} & \textbf{95.8} && 94.1 & 74.6 && 89.7 & 73.0 & 94.2 & 75.5 && 95.6 & 84.6 \\
7.0 && \textbf{98.9} & \underline{96.8} & 97.7 & 91.3 & \underline{98.9} & 94.7 && 92.8 & 70.0 && 90.5 & 73.3 & 94.2 & 76.0 && 95.5 & 83.7 \\
\midrule
\midrule
\rowcolor{gray!15}
Adaptive (ours) && \textbf{99.0} & \textbf{97.0} & \underline{97.9} & \underline{92.6} & \underline{99.0} & \underline{95.2} && \textbf{96.6} & \textbf{83.7} && \underline{93.6} & \textbf{81.3} & \underline{96.7} & \underline{83.6} && \textbf{97.1} & \textbf{88.9} \\
\midrule
\end{tabular}%
}
\end{table}

\vspace{5pt} \noindent \textbf{AGS Component Ablation} \\
Table~\ref{tab:ags_components} evaluates three components of AGS in the few-shot setting. BFP denotes boundary-filtered prototype computation. When BFP is disabled, class prototypes are computed using all points; when enabled, points near organ transitions are removed before prototype computation. $S_{\mathrm{boundary}}^{g,c}$ denotes the boundary-aware granularity score. When disabled, granularity selection uses only the internal score based on inter-class separation and intra-class compactness; when enabled, separate class-wise boundary scores are also used. CWA denotes confidence-weighted aggregation, where the selected granularity predictions are combined using the average softmax confidence of the MLP.
The results show that the components are most effective when used together. BFP alone does not improve performance, because boundary points are removed from prototype computation but are not explicitly evaluated during granularity selection. Adding $S_{\mathrm{boundary}}^{g,c}$ improves the average mIoU, and combining it with BFP improves it further. The full AGS pipeline, using BFP, $S_{\mathrm{boundary}}^{g,c}$, and CWA, achieves the best average mIoU of 88.9\%.

\begin{table}[t]
\centering
\caption{\textbf{Ablation of AGS components in the few-shot setting.}
BFP denotes boundary-filtered prototype computation.
$S_{\mathrm{boundary}}^{g,c}$ indicates whether boundary-aware granularity scores are used.
CWA denotes confidence-weighted aggregation at inference time.
Best result is shown in bold and second-best is underlined.}
\label{tab:ags_components}
\begin{tabular}{lcccccccccc}
\toprule
Method &&& BFP && $S_{\mathrm{boundary}}^{g,c}$ && CWA && Avg. mIoU \\
\midrule
Ablation 1 &&&  &&  &&  && 87.3 \\
Ablation 2 &&& \cmark &&     &&  && 86.8 \\
Ablation 3 &&&  && \cmark &&  && 88.3 \\
Ablation 4 &&& \cmark    && \cmark    &&  && \underline{88.5} \\
\midrule
\rowcolor{gray!15}
Ours       &&& \cmark    && \cmark    && \cmark    && \textbf{88.9} \\
\bottomrule
\end{tabular}
\end{table}

\subsection{Runtime and Memory Footprint}
We also report the runtime and memory footprint of our AGS module. In our current implementation, AGS requires roughly 15\,s per plant during both training and inference.
This includes feature extraction and granularity selection, and could likely be reduced through a more efficient search strategy and parallelization. 
In the full-split setting, AGS takes approximately 6\,min, followed by 15\,min for MLP training, resulting in a total training time of about 21\,min. In the few-shot setting, AGS takes approximately 45\,s and MLP training takes about 10\,min, for a total of roughly 11\,min (resulting in a $2\times$ relative speedup). The one-shot setting is substantially cheaper, requiring approximately 3.5\,min in total (resulting in a $6\times$ relative speedup). 
The main practical difference is memory usage. In our implementation, the full-split setting exceeded the available memory on an RTX 5090, requiring heavy subsampling of the training features. In contrast, the few-shot and one-shot settings use fewer annotated plants and therefore have a much smaller memory footprint.

\section{Conclusion}

We propose AGS-PlantSeg, a few-shot 3D plant organ segmentation method that leverages the frozen Utonia model through Adaptive Granularity Selection. 
Instead of relying on a single fixed granularity, AGS selects suitable granularity levels for each plant using internal class separability and boundary-aware scores, and applies them with a lightweight MLP segmentation head. 
The results show that the best fixed granularity level varies across species and datasets. AGS improves cross-species robustness and achieves the best average performance, reaching 88.9\% mIoU, and outperforming the best fixed-granularity baseline by 2.5 mIoU points.
The few-shot model remains close to the full-split setting while requiring substantially fewer annotated plants, suggesting that frozen 3D foundation features can reduce annotation requirements for plant organ segmentation. \\

\noindent \textbf{Limitations and Future Work.}
Although AGS is not always optimal for every individual species, its strong average performance shows that adaptive granularity selection is a promising direction for generalizable 3D plant phenotyping. Future work could investigate bigger segmentation heads, stronger training-time augmentation, and more efficient granularity search and scoring strategies. AGS could also be extended to a broader range of plant morphologies, including crops such as wheat and rice, and to additional semantic classes, such as fruits and soil.
This work focuses specifically on evaluating whether adaptive granularity selection improves generalization across plant species and supervision settings. Future work could include a broader comparison with state-of-the-art few-shot segmentation approaches, particularly prototype-based methods such as COSeg \cite{coseg2024}, as well as an analysis of how the selected granularity levels vary with species, plant morphology, and growth stage.

\par\vfill\par

\section*{Acknowledgements}

This work has been supported by the Novo Nordisk Foundation under grant agreement No. NNF25SA0104538 (Robotic Intercropping).

%
%
\bibliographystyle{splncs04}
\bibliography{main}
\end{document}